# SS3D: Single Shot 3D Object Detector

Aniket Limaye*[1]*, Manu Mathew, Soyeb Nagori, Pramod Kumar Swami, Debapriya Maji, Kumar Desappan

Texas Instruments (India) Ltd.

## Abstract

*Predicting accurate 3D bounding boxes of objects from the given scene is important for autonomous driving [1]. 3D object detection typically uses LiDAR point cloud or image(s) input to do the prediction. The algorithm used must be simple enough to be able to run in real time on embedded systems without compromising accuracy of detections. Single stage deep learning algorithm for 2D object detection was made popular by Single Shot MultiBox Detector (SSD) [13] and it was heavily adopted in several embedded applications. PointPillars [1] is a fast 3D object detection algorihm that produces state of the art results and uses SSD adapted for 3D object detection. The main downside of PointPillars is that it has a two stage approach with learned input representation based on fully connected layers followed by SSD.*

*[1] In this paper we present Single Shot 3D Object Detection (SS3D) - a single stage 3D object detection algorithm which combines a straight forward, statistically computed input representation and a single shot object detector based on PointPillars. This can be considered as a single shot deep learning algorithm as computing the input representation is straight forward and does not involve much computational cost. We show that the proposed SS3D algorithm can be just as effective as the popular two stage detectors when using LiDAR point cloud as input. Due to the use of a single stage approach, our proposed method is suitable for embedded applications. We also extend our method to stereo input and show that, aided by additional semantic segmentation input; our method produces similar accuracy as state of the art stereo based detectors.*

*Achieving the accuracy of two stage detectors using a single stage approach is a important for 3D object detection as single stage approaches are simpler to implement in real-life applications. With LiDAR as well as stereo input, our method outperforms PointPillars, which is one of the state-of-the art methods for 3D object detection. When using LiDAR input, our input representation is able to improve the $AP_{3D}$ of Cars objects in the moderate category from **74.99** to **76.84**. When using stereo input, our input representation is able to improve the $AP_{3D}$ of Cars objects in the moderate category from **38.13** to **45.13**. Our results are also better than other popular 3D object detectors such as AVOD [7] and F-PointNet [8].*

---

[1] Aniket Limaye is pursuing his masters as IIT Bombay. This research was done while he was an intern at Texas Instruments (India) Ltd, Bangalore.

## 1. Introduction

Several algorithms to do 2D object detection from images have become popular in the recent years. To make autonomous vehicles feasible, object detection has to be done in 3D to enable trajectory planning in bird's eye view (BEV) representation. The best results in 3D object detection so far have been obtained by using LiDAR (Light Detection and Ranging) point clouds as inputs [1]. LiDAR gives accurate measurements in 3D which helps to get high accuracy in 3D object detection.

2D object detection can be either done as a two stage detector as in Faster-RCNN [12] or in as a single stage detector as in Single Shot MultiBox Detector (SSD) [13]. The two stage approaches are computationally expensive and do not give good frames per second (FPS) performance on embedded systems. Hence single stage methods are highly desirable in embedded systems.

Several popular 3D object detection algorithms use two stage approaches to extract features and/or to do the actual detection. In some methods, the core detector itself has multiple stages, such as in AVOD [7] and F-PointNet [8]. In other methods such as VoxelNet [9], SECOND [10] and PointPillars [1] the input representation involves using a separate stage of learned neural network layers to generate learned features before the actual detector. These multi-stage approaches make the overall application slower and difficult to implement on embedded systems.

In this work we propose Single Shot 3D Object Detection (**SS3D**), an algorithm that has a simple to compute input representation followed by a Single Shot Detector. This can be considered as a single stage method, since the computation of input representation is light weight. Thus a light-weight input representation can be given to our fast detector that performs 3D object detection in a single shot manner.

We also extend our work to take stereo input – in particular we use the method outlined in Pseudo-LiDAR [2] to first convert the stereo disparity to a point cloud representation before applying SS3D. In the case of stereo, we also found that it is beneficial to provide 2D semantic segmentation information computed from the input image as an additional input to improve the accuracy further. Our extended method for stereo is called **SS3D-Seg.** This stereo method cannot be considered single stage as it involves stereo disparity computation and/or semantic segmentation – however, the core detector is still the single stage SSD.

We benchmark the proposed algorithms on the KITTI

| Method Name | Input representation |
|---|---|
| PointPillars [1] | 100 points transformed into 64 features by a fully connected layer. |
| ApolloAuto [3][4] | 8 values per pillar (*Occupied bit*, Number of points in pillar, Mean height of points, Mean intensity of points, Max height of points in pillar, Intensity of the highest point in pillar, Distance of pillar center from the origin/camera, Angle of pillar center with respect to the origin) |
| SS3D-6 | 6 values per pillar (*Occupied bit*, Number of points in pillar, Mean height of points, Mean intensity of points, Max height of points in pillar, Intensity of the highest point in pillar) |
| SS3D-10 | 10 values per pillar (Number of points in pillar, Mean height of points, Mean intensity of points, Max height of points in pillar, Intensity of the highest point in pillar, Distance of pillar center from the origin/camera, Angle of pillar center with respect to the origin, **3 max heights below different horizontal planes**) |
| SS3D-Seg-6 | 6 values per pillar (*Occupied bit*, Number of points in pillar, Mean height of points, **Mean segmentation mask value** of points – which is the average of the label values in the pillar, Max height of points in pillar, **Segmentation Mask value** of the highest point in pillar) |
| SS3D-Seg-10 | 10 values per pillar (Number of points in pillar, Mean height of points, **Mean segmentation mask value** of points, Max height of points in pillar, **Segmentation mask value** of the highest point in pillar, Distance of pillar center from the origin/camera, Angle of pillar center with respect to the origin, **3 max heights below different horizontal planes**) |

Table 1: Details of the input representations

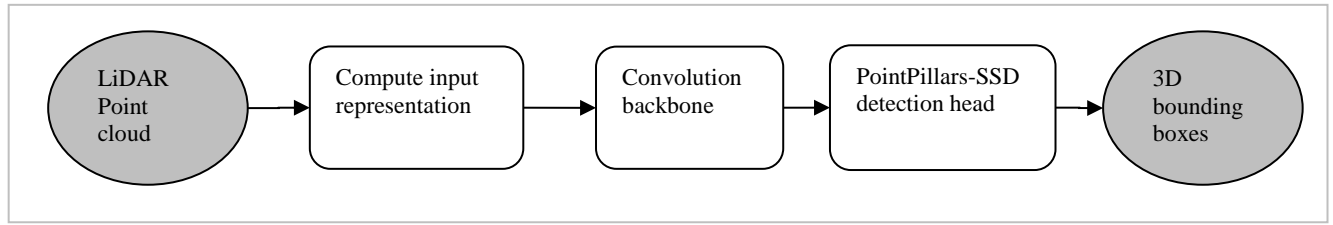

Figure 1. Block diagram of the proposed Single Shot 3D Object Detector (SS3D) with LiDAR point cloud input

3D Object Detection Evaluation 2017 [5][6] benchmark and show competitive accuracy compared to state of the art methods on LiDAR input as well as on stereo input. Being able to use a single stage method instead of two stage method will help to simplify the implementation in real-life applications.

## 2. Related Work

An exhaustive review of 3D object detection algorithms is given in PointPillars [1]. Here we only review a few state of the art algorithms that are important for this work.

- Viola & Jones face detector was one of the most successful detector and it used features computed using Haar basis functions and AdaBoost classification.
- Histogram of oriented gradients (HOG) [21] was another popular object detector and that also used classical computer vision and machine learning techniques such as hand-crafted features and Support Vector Machines (SVM) for classification.
- The series of pascal visual object classes challenges (VOC) [16] was one of the most popular 2D object detection challenges to date and it helped researches to benchmark algorithms quickly and effectively.
- During this time the ImageNet classification dataset [17] and ILSVRC [18] benchmark became effective methods to measure the effectiveness of image classifiers and also Convolutional Neural Networks (CNN).
- Faster-RCNN [12] was one the earliest CNN based object detectors to produce high accuracy predictions at reasonable complexity. However it uses two stages
- in its network separated by Region
- of Interest Pooling (ROI Pooling). ROI pooling operation is data bandwidth intensive in nature, reducing the possibility of its use in low power embedded systems.
- Single Shot MultiBox Detector (SSD) [13] is one of the most popular single stage object detectors that achieve high accuracy detections comparable to that of Faster-RCNN without having expensive ROI Pooling operations. It has high degree of adoption into embedded systems, providing us with a guideline on how to design high throughput object detectors for low power embedded systems.

- The KITTI Vision Benchmark Suite [5] and its KITTI 3D Object Detection Evaluation 2017[6] is one of the most popular benchmark to compare 3D object detection algorithms.

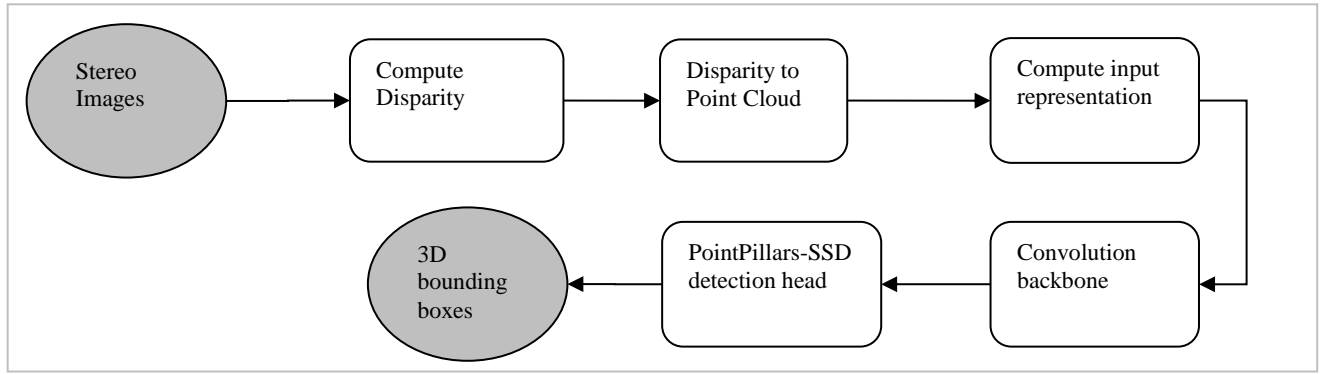

**Figure 2 Block diagram of the proposed SS3D with Stereo Images input**

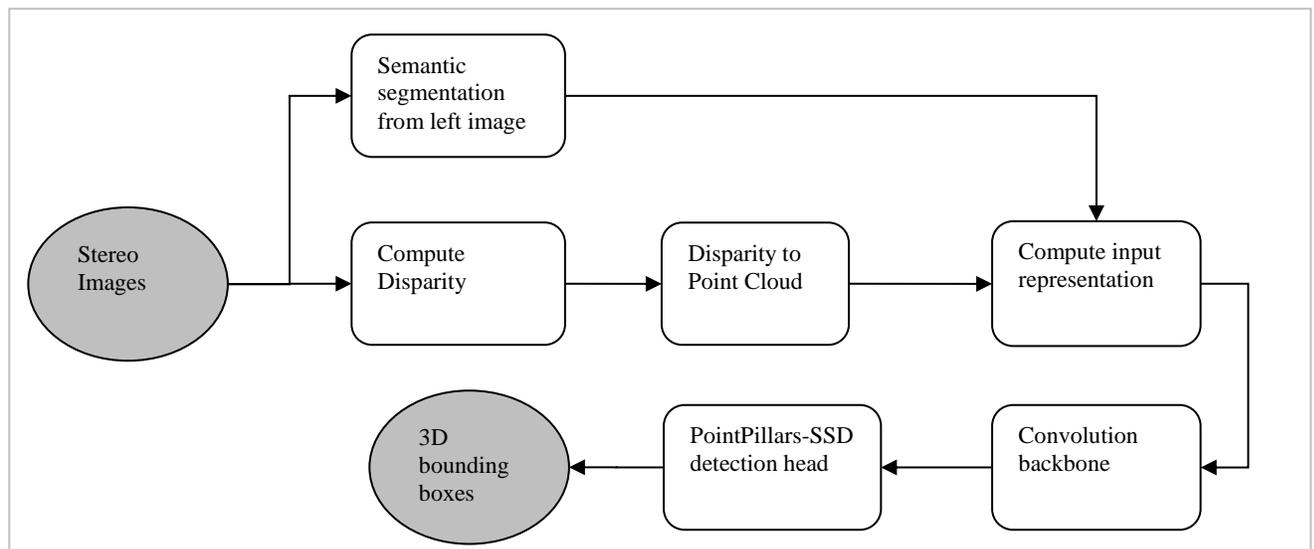

**Figure 3 Block diagram of the proposed SS3D-Seg with Stereo Images input**

- AVOD [3] and F-PointNet [8] are state of the art 3D object detection algorithms that provides high accuracy on LiDAR combined with image input. However these use multi stage approaches similar to Faster-RCNN. Multi stage approaches involving operations such as ROI pooling are data bandwidth intensive in nature and are not suitable for embedded implementation.
- The accuracy of stereo based 3D object detectors used to be quite poor. Pseudo-LiDAR [2] showed that if the stereo disparity is first converted to a point cloud and then point cloud based 3D object detectors are applied, then much higher accuracies can be achieved. It applied this method on AVOD [3] and F-PointNet [8] showed much higher accuracies compared to existing stereo based detectors. For example the $AP_{3D}$ (3D Average Precision) at 0.7 Intersection over Union (IoU) for Cars in the KITTI 3D objection detection benchmark increased from just 6.6% for the basic AVOD method to 45.3% for Pseudo-LiDAR with AVOD.
- PointPillars [1] is yet another 3D object detector that provides quite high accuracy with LiDAR input. One interesting thing about PointPillars is a clear separation between input representation and the core detector. The core detector is a single stage detector – unlike AVOD and F-PointNet. However computing the input representation is quite complex as it involves several fully connected operations as in a fully connected neural network layer. The utilization of the compute units in processors (GPU or CPU) may also be poor as these fully connected operations need to be done separately for each pillar in the PointPillars grid of pillars. To take an example of compute complexity, for the 496x432 image resolution that is used as default in PointPillars, the fully connected operations alone contribute 691 Mega Multiply Accumulations (typically referred to as Mega Flops in Deep Learning literature [28]).

| Config | | | | AP$_{BEV}$ | | | AP$_{3D}$ | | |
|---|---|---|---|---|---|---|---|---|---|
| Input | Method Name | Input representation | Detector | Easy | Mod | Hard | Easy | Mod | Hard |
| LiDAR | AVOD | Psuedo-LiDAR [2] | AVOD | 88.53 | 83.79 | 77.9 | 81.94 | 71.88 | 66.38 |
| LiDAR | FPointNet | Psuedo-LiDAR [2] | F-PointNet | 88.7 | 84 | 75.33 | 81.2 | 70.39 | 62.19 |
| LiDAR | PointPillars | PointPillars | PointPillarsSSD | 88.35 | 86.1 | 79.83 | 79.05 | 74.99 | 68.3 |
| LiDAR | SS3D-6 | SS3D-6 | PointPillarsSSD | **90.18** | **87.32** | **84.97** | **87.19** | **76.84** | **70.16** |
| LiDAR | SS3D-10 | SS3D(10) | PointPillarsSSD | 90.07 | 86.39 | 83.99 | 86.79 | 76.48 | 69.98 |

Table 2: Results with LiDAR input for Car detection in the KITTI 3D OD evaluation. *(Results in Blue color are generated by us).*

| Config | | | | AP$_{BEV}$ | | | AP$_{3D}$ | | |
|---|---|---|---|---|---|---|---|---|---|
| Input | Method Name | Input representation | Detector | Easy | Mod | Hard | Easy | Mod | Hard |
| Stereo | 3DOP [24][2] | 3DOP | 3DOP | 12.6 | 9.5 | 7.6 | 6.6 | 5.1 | 4.1 |
| Stereo | MLF-STEREO [25][2] | MLF-STEREO | MLF-STEREO | - | 19.5 | - | - | 9.8 | - |
| Stereo | AVOD | Psuedo-LiDAR [2] | AVOD | 74.9 | **56.8** | **49** | 61.9 | **45.3** | **39** |
| Stereo | F-PointNet | Psuedo-LiDAR [2] | F-PointNet | 72.8 | 51.8 | 44 | 59.4 | 39.8 | 33.5 |
| Stereo | PointPillars | Psuedo-LiDAR | PointPillarsSSD | 68.71 | 49.45 | 43.82 | 56.08 | 38.13 | 32.61 |
| Stereo | SS3D-Seg-10 | SS3D-Seg-10 | PointPillarsSSD | **76.97** | 55.61 | 48.28 | **65.42** | 45.13 | 38.18 |

Table 3: Results with Stereo input for Car detection in the KITTI 3D OD evaluation. *(Results in Blue color are generated by us).*

- ApolloScape [3] is a large open dataset for autonomous driving and ApolloAuto is an open autonomous driving platform based on ApolloScape. ApolloAuto 3D object detection [4] uses statistically computed features and a fully convolutional network to come up with an intermediate representation for cells in the bird's eye view plane. This intermediate representation is then sent to a post processing stage to obtain the predictions about obstacles.

## 3. Single Shot 3D Object Detector (SS3D)

The overall detector is inspired by PointPillars [1] 3D object detection method. It uses fully connected layers to compute an input representation which is fed to an SSD object detection backend. These fully connected layers are computationally expensive to compute. In the proposed method, we do not use the input representation used in PointPillars that involves fully connected layers. Instead, we use a lightweight input representation inspired by ApolloAuto [3].

### 3.1. Input representation

The range of point cloud is capped at about 80m wide and 70m deep in the horizontal plane and -3m to 1m in height with respect to the LiDAR. This region is divided into a grid of size 496x432 where each grid box has a spatial size of 0.16m x 0.16m x 4m. All the points falling into a grid box are used to generate the features representing that box. This can be thought of as a pillar with the dimensions of the grid-box and the height chosen (0.16m x 0.16m x 4m). The input representation for each pillar is extracted from the 3D points (from the point cloud) that fall within that pillar. Depending on the details of input representation, there are a few variants of the proposed approach as given below. These input representations are summarized in Table 1.

- **Point cloud based representation:** In this method a set of representative values are computed per pillar. In we denote this as SS3D-6, since we compute the following 6 values: (1) Occupied bit indicating whether the current pillar has at-least one valid point or not, (2) Number of points in the pillar, (3) Mean height of points in the pillar, (4) Mean intensity of points in the pillar (mean LiDAR reflectivity in the case of LiDAR input), (5) Max height of points in the pillar, (6) Intensity of the highest point in the pillar. Thus the size of the input representation is 496x432x6. This variant is called SS3D-6. In another variant we slice the height into 3 parts and take the maximum height value within each slice of each pillar. We also use the angle of the pillar center with respect to the origin. Thus we get 4 additional values in this input representation – 3 maximum heights and one angle per pillar – forming a total of 10 values - and we call this representation SS3D-10. The block diagram of this approach is given in Figure 1.

| Config | | | | AP$_{BEV}$ | | | AP$_{3D}$ | | |
|---|---|---|---|---|---|---|---|---|---|
| Input | Method Name | Input representation | Detector | Easy | Mod | Hard | Easy | Mod | Hard |
| Stereo | PointPillars | PointPillars | PointPillarsSSD | 68.71 | 49.45 | 43.82 | 56.08 | 38.13 | 32.61 |
| Stereo | SS3D-6 | SS3D-6 | PointPillarsSSD | 74.26 | 50.59 | 44.2 | 53.71 | 37 | 30.59 |
| Stereo | SS3D-Seg-6 | SS3D-Seg-6 | PointPillarsSSD | 76.87 | 55.1 | **48.58** | 61.91 | 40.33 | 36.89 |
| Stereo | SS3D-Seg-10 | SS3D-Seg-10 | PointPillarsSSD | **76.97** | **55.61** | 48.28 | **65.42** | **45.13** | **38.18** |

**Table 4: Ablation study - comparison of the input representations for Car detection in the KITTI 3D OD evaluation.**
*(Results in Blue color are generated by us).*

- **Point cloud and image based representation**: In the case of the stereo input, we can use the point cloud generated from the stereo images as shown by Pseudo-LiDAR [2]. Thus we can create the SS3D-6 and SS3D-10 input representations from stereo input as well. The block diagram of this approach is given in Figure 2. In addition we can generate semantic segmentation information from the left image and provide as an additional input while creating the input representation. We call these input representations as SS3D-Seg-6 and SS3D-Seg-10. The block diagram of this method is given in Figure 3.

Details of these approaches are summarized in Table 1.

### 3.2. Convolutional backbone

The convolutional backbone used in the proposed approach is same as in PointPillars. It consists of two sub-networks. The first consists of a series of blocks that reduce the spatial resolution of the input feature maps. Each block consists of a series of convolutional layers followed by Batch-Norm and a ReLU. The second network takes the features from each block as input. These multi-level features are up-sampled to a common spatial resolution through a transposed 2D convolution each with the same number of output features. These final features coming from different spatial resolutions are then concatenated to form the final feature map.

### 3.3. Detection head

We use the same SSD detection head that is used by PointPillars. 2D Inter-section over Union (IoU) is used for matching the prior-boxes to the ground truth during training. The detection head predicts the location on the horizontal plane. Bounding box height and elevation were predicted as additional regression targets.

## 4. Implementation Details

Our method was implemented on the PointPillars code base and used the same hyper parameters whenever possible. We trained the model on 2 types of inputs: LiDAR, and Stereo images.

• LiDAR point clouds were directly obtained from LiDAR sensor and we pass this on to the creation of input representation and the SSD model.

• For Stereo input, we generate the disparity map using PSMNET [23] trained on the KITTI [5] training images as done in [2].

PointPillars use the reflectivity information from the LiDAR to compute the mean intensity in the pillars. However, in the case of Pseudo-LiDAR, these values are populated as 1's in the input representation, wherever it is mentioned. We follow the same convention as well in our proposal. We tried to replace those fixed values with the grayscale intensity values from image, however it did not produce improved results and hence it is not part of our proposal.

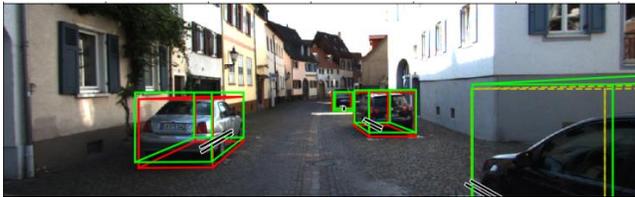
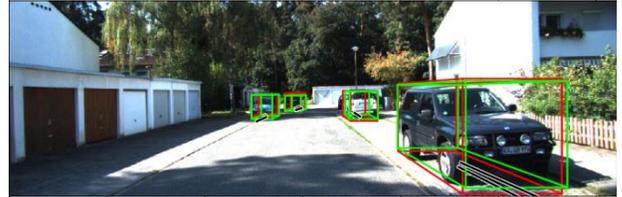

**Figure 4. Sample detections with LiDAR input and SS3D-Seg-6 input representation. Green boxes are the detections and red boxes are the ground truth**

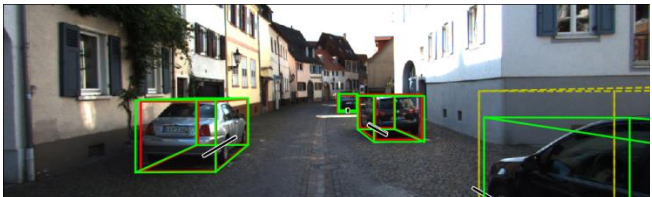

**Figure 5. Sample detections with Stereo input and SS3D-Seg-10 input representation. Green boxes are the detections and red/yellow boxes are the ground truth.**

## 5. Experiments and Results

We train and benchmark the methods on KITTI 3D Object Detection [6] dataset and benchmark. The training a validation splits is same as the ones used by PointPillars. All the results reported are for the **Car detection task** and the IoU threshold used is 0.7 as per the benchmark.

The baseline results for LiDAR are those that are published from PointPillars [1] and for Stereo are those from Pseudo-LiDAR [2]. In our proposed methods, we remove the input representation in PointPillars and replace it with the proposed ones.

Table 2 shows the results for LiDAR input. We can see that the proposed SS3D-6 input representation coupled with PointPillarsSSD outperforms all other methods that we compared against in terms of accuracy, demonstrating that the proposed input representation is as powerful as a learned representation for 3D object detection task. Specifically, when using LiDAR input, our input representation is able to improve the $AP_{3D}$ of Cars objects in the moderate category from **74.99** to **76.84**.

Next, we applied the proposed method to Stereo task. We first generated disparity using PSMNET and then converted it to point cloud as suggested by Pseudo-LiDAR. We also generate semantic segmentation predictions on the left image, wherever needed. The semantic segmentation was generated using a variant of the DeepLabV3+ [26] architecture with a MobileNetV2 backbone, trained on the Cityscapes dataset [27]. This information is used to create the input representation. Table 3 compares the results of our method with those in the literature. It is seen that the proposed method achieves competitive accuracy compared to the state of the art methods using stereo input. Specifically, when using stereo input, our input representation is able to improve the $AP_{3D}$ of Cars objects in the moderate category from **38.13** to **45.13**.

With LiDAR as well as stereo input, our method outperforms PointPillars, which is one of the state-of-the art and fast method for 3D object detection.

## 6. Ablation study

Table 4 shows an ablation study for stereo input about the effectiveness of the various values used in the input representation. We find that the semantic segmentation information is especially useful in improving the accuracy. The extra 3 values based on height slicing were also helpful in improving the accuracy.

## 7. Conclusion

We present a single stage 3D object detection method for point clouds, especially for LiDAR input. Our results show that the proposed method has accuracy comparable to that of state of the art methods. We also adapt our method for stereo input and show that we can achieve competitive 3D object detection accuracy using an SSD-like object detector. In the case of LiDAR as well as with stereo input, the results are better compared to that of original PointPillars, showing that these input representations are sufficient and powerful enough for 3D object detection task. We believe that our finding will enable 3D object detection in low cost embedded applications where single shot methods have a throughput advantage.

## 8. Acknowledgements

We thank the authors of PointPillars [1] and Pseudo-LiDAR [2] for making their code base publicly available. This helped us to easily implement our proposed methods on top of it.